\pdfoutput=1

\documentclass[11pt]{article}

\usepackage[preprint]{acl}

\usepackage{times}
\usepackage{latexsym}

\usepackage[T1]{fontenc}

\usepackage[utf8]{inputenc}

\usepackage{microtype}

\usepackage{inconsolata}

\usepackage{graphicx}

\usepackage{algorithm}
\usepackage{algorithmic}
\usepackage{amsmath}
\usepackage{booktabs}
\usepackage{multirow}
\usepackage{subfigure}

\title{Scaffold-BPE: Enhancing Byte Pair Encoding for Large Language Models with Simple and Effective Scaffold Token Removal}

\author{Haoran Lian \\
  Beihang University \\
  \texttt{email@domain} \\\And
  Second Author \\
  Affiliation / Address line 1 \\
  Affiliation / Address line 2 \\
  Affiliation / Address line 3 \\
  \texttt{email@domain} \\}

\author{
 \textbf{Haoran Lian\textsuperscript{1}},
 \textbf{Yizhe Xiong\textsuperscript{2,3}},
 \textbf{Jianwei Niu\textsuperscript{1}},
 \textbf{Shasha Mo\textsuperscript{1}},
 \textbf{Zhenpeng Su\textsuperscript{4}},\\
 \textbf{Zijia Lin\textsuperscript{2}},
 \textbf{Hui Chen\textsuperscript{2,3}},
 \textbf{Peng Liu\textsuperscript{5}},
 \textbf{Jungong Han\textsuperscript{2}},
 \textbf{Guiguang Ding\textsuperscript{2,3}},
\\
\\
 \textsuperscript{1}Beihang University,
 \textsuperscript{2}Tsinghua University,
 \textsuperscript{3}BNRist,\\
 \textsuperscript{4}Chinese Academy of Sciences,
 \textsuperscript{5}Beijing Institute Of Technology
\\
 \small{
   \textbf{Correspondence:} \href{mailto:niujianwei@buaa.edu.cn}{niujianwei@buaa.edu.cn}
 }
}

\begin{document}
\maketitle
\begin{abstract}
Byte Pair Encoding (BPE) serves as a foundation method for text tokenization in the Natural Language Processing (NLP) field. Despite its wide adoption, the original BPE algorithm harbors an inherent flaw: it inadvertently introduces a frequency imbalance for tokens in the text corpus. Since BPE iteratively merges the most frequent token pair in the text corpus to generate a new token and keeps all generated tokens in the vocabulary, it unavoidably holds tokens that primarily act as components of a longer token and appear infrequently on their own. We term such tokens as \textbf{Scaffold Tokens}. Due to their infrequent occurrences in the text corpus, Scaffold Tokens pose a learning imbalance issue. To address that issue, we propose \textbf{Scaffold-BPE}, which incorporates a dynamic scaffold token removal mechanism by parameter-free, computation-light, and easy-to-implement modifications to the original BPE method. This novel approach ensures the exclusion of low-frequency Scaffold Tokens from the token representations for given texts, thereby mitigating the issue of frequency imbalance and facilitating model training. On extensive experiments across language modeling and even machine translation, Scaffold-BPE consistently outperforms the original BPE, well demonstrating its effectiveness.
\end{abstract}

\section{Introduction}


In recent years, Large Language Models (LLMs) have become a burgeoning paradigm in handling a broad array of Natural Language Processing (NLP) tasks. The tokenization process in most modern LLMs \cite{radford2019language,brown2020language,rae2021scaling,zhang2022opt,biderman2023pythia,touvron2023llama,yang2023baichuan,achiam2023gpt,dubey2024llama} employs Byte Pair Encoding (BPE) \cite{sennrich2015neural}, a method that was originally designed for data compression \citep{gage1994new}. BPE consists of two main stages. In the training stage, BPE iteratively merges the most frequent pairs of bytes or characters in a dataset into a new token and adds it to the vocabulary until a desired vocabulary size is reached. And in the encoding stage, the vocabulary is utilized to represent any text. The adoption of BPE in LLMs is driven by its capability to decompose words into smaller, manageable subword units, thus avoiding out-of-vocabulary words, facilitating flexible and semantically complete representations of input data. 
Actually, BPE has also been widely used in traditional NLP tasks, like machine translation \citep{provilkov2019bpe,xu2020vocabulary}, sentence classification \citep{liu2019roberta,he2020deberta} and summarization \citep{wu2021bass,xu2022sequence}.

\begin{figure}[t]
\centering
\includegraphics[width=0.95\columnwidth]{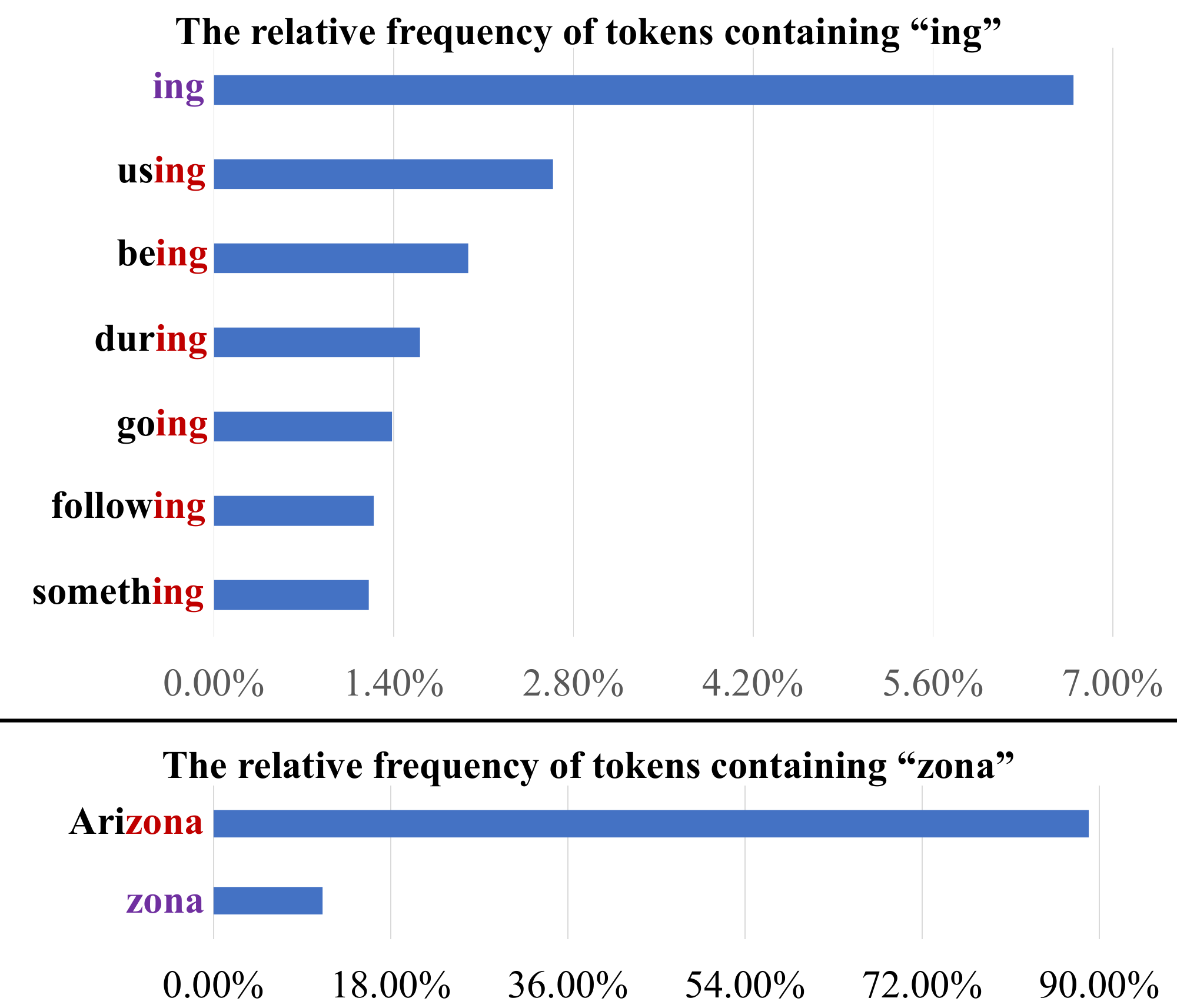} 
\caption{Two types of tokens in original BPE vocabulary on the Pile dataset: one type, such as ``ing'', appears frequently by itself, while the other type, such as ``zona'', mostly appears as a component of ``Arizona'' and thus has a low individual occurrence frequency. The value on the horizontal axis represents the percentage of the frequency of the token relative to the total frequency of all tokens containing ``ing''/``zona''. For ``ing'', we only visualize the top tokens.}
\label{Case Study}
\end{figure}

Since its inception, BPE has undergone various modifications to better suit the needs of complex NLP tasks, including identifying the optimal vocabulary size for various tasks \cite{xu2020vocabulary,gutierrez2021characters}, optimizing the encoding paths of tokens to achieve subword regularization \cite{provilkov2019bpe,he2020dynamic}, etc.

\begin{figure}[t]
\centering
\includegraphics[width=0.95\columnwidth]{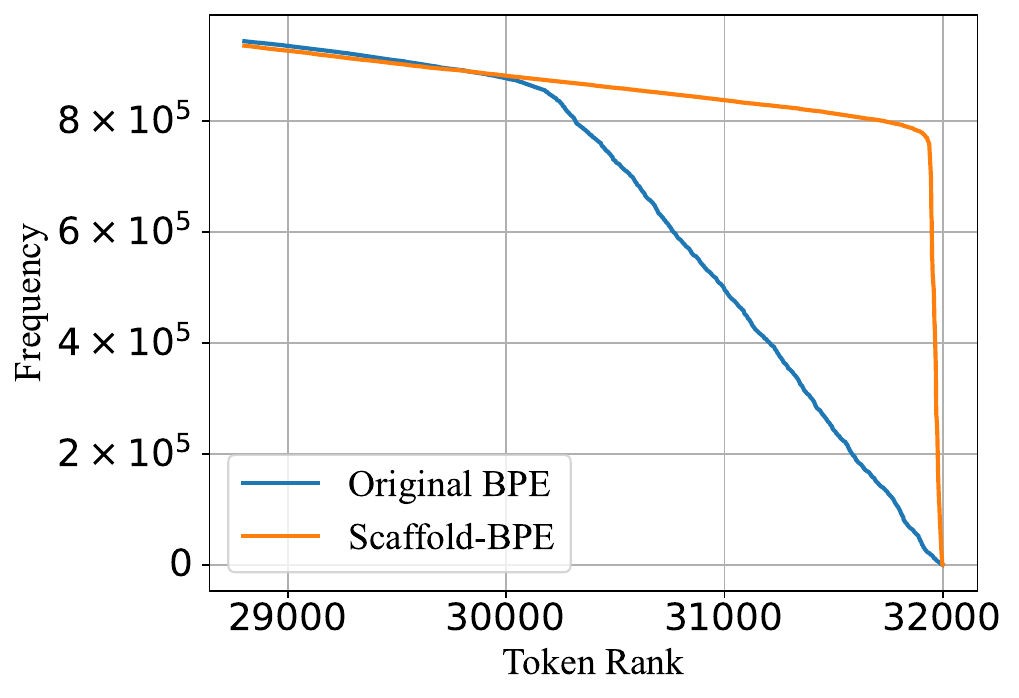}
\caption{Sorted token frequencies in descending order of the original BPE and Scaffold-BPE.}
\label{token_freq_diff}
\end{figure}

However, existing studies have overlooked an inherent limitation in the BPE method: the iterative merging process can lead to an imbalance in token frequencies by including low-frequency tokens in vocabulary. For example, as illustrated in Figure \ref{Case Study}, in the commonly used Pile dataset \citep{gao2020pile} that is tokenized by the original BPE method of 32K vocabulary size (as LLaMA series \cite{touvron2023llama,touvron2023llama2}), the token ``zona'' mostly appears as a component of the token ``Arizona'' rather than as an independent, high-frequency token. Despite its lower standalone frequency, BPE includes ``zona'' in the final vocabulary because it is the ``intermediate token'' to derive the frequent token ``Arizona''. We define such intermediate tokens that are crucial for constructing longer frequent tokens but do not appear frequently on their own as \textbf{Scaffold Tokens}.
Note that not all subwords are simply scaffold tokens. For example, ``ing" is not identified as a scaffold token, as there are many words containing ``ing" but are not tokens in the vocabulary. For example, ``connecting" is represented as ``connect"+``ing" (2 tokens). Such words help to keep “ing” a frequent token. Therefore, ``ing" is not a scaffold token. 
According to our proposed Scaffold-BPE method, the 32K vocabulary contains about 6.07\% of scaffold tokens.

As depicted in Figure \ref{token_freq_diff}, a natural frequency imbalance arises between these scaffold tokens and actual high-frequency tokens. Prior studies \cite{lin2017focal,su2023infoentropy} have highlighted that such disparities in token frequencies can result in imbalanced learning difficulties across different tokens. Scaffold tokens, due to their lower individual occurrence frequencies, are notably harder to learn for models. 

To address that issue, we propose enhancements to the BPE algorithm, aiming to mitigate the frequency imbalance and ensure a more equitable learning process for all tokens.
Specifically, we propose the simple and effective \textbf{Scaffold-BPE} with a dynamic scaffold token removal mechanism, which is parameter-free, computation-light, easy-to-implement, and widely effective. 
Generally, the proposed Scaffold-BPE maintains an expanded vocabulary compared with the original BPE, which consists of both normal tokens and scaffold tokens. Note that the scaffold tokens are not actual tokens in the vocabulary and do not appear in the tokenized sequences after encoding. 
In the training stage, Scaffold-BPE dynamically marks tokens with lower individual occurrence frequencies as scaffold tokens in each iteration. In the encoding stage, the Scaffold-BPE firstly utilizes all tokens in the expanded vocabulary to generate the token representations for the given texts, which is termed as a \textit{Scaffolding} process. Then, the Scaffold-BPE ensures the absence of all scaffold tokens in the token representation by demolishing them into their shortest non-scaffold-token sequence, which is termed as a \textit{Demolishing} process. Thanks to such modifications, Scaffold-BPE can remove scaffold tokens from the final token representations fed into models, thus enjoying more balanced token occurrences, leading to more sufficient learning and better performance of models.

We conduct extensive experiments on language modeling tasks. Results on 9 widely used language modeling benchmarks demonstrate that Scaffold-BPE consistently outperforms the original BPE. Besides, even when extended to machine translation tasks, Scaffold-BPE proves highly effective. Furthermore, we show that Scaffold-BPE is orthogonal to existing modifications on BPE, like BPE-Dropout \cite{provilkov2019bpe} and can be combined with them to achieve further improvements.

\begin{figure*}[t]
\centering
\includegraphics[width=0.95\textwidth]{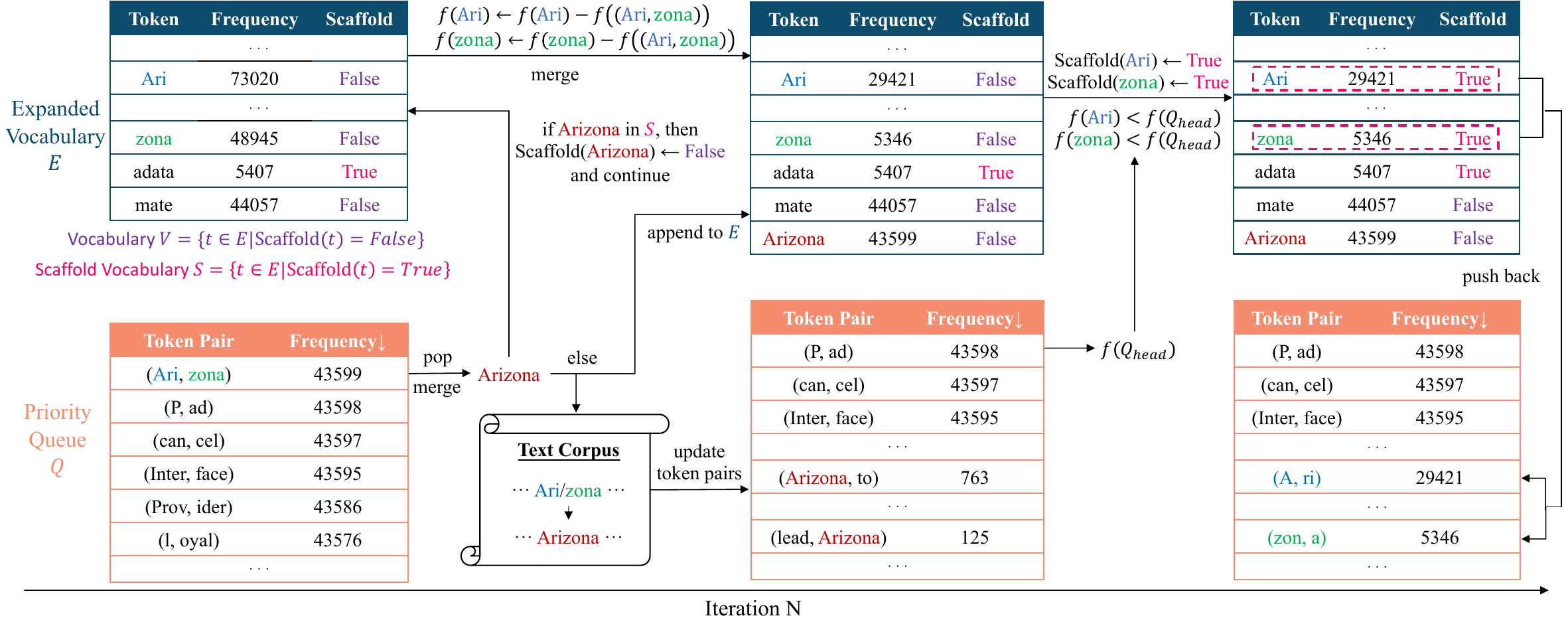}
\caption{Illustration of one iteration in the Scaffold-BPE training process.}
\label{pipeline}
\end{figure*}

Overall, our contributions are three-fold:
\begin{itemize}
    \item We observe that the iterative training process of BPE incorporates low-frequency tokens into the vocabulary, which we term scaffold tokens.
    \item We propose Scaffold-BPE, which can remove scaffold tokens from the final token representations by dynamically marking scaffold tokens in the training process and temporarily utilizing scaffold tokens in the encoding process. Scaffold-BPE is parameter-free, computation-light, easy-to-implement, and widely effective, preserving the simplicity and clarity of BPE.
    \item Extensive experiments demonstrate that Scaffold-BPE outperforms the original BPE on language modeling and also machine translation tasks, proving its effectiveness and robustness in the NLP field.
\end{itemize}

\section{Related Works}

Recently, LLMs have become a popular paradigm for solving NLP tasks, with BPE serving as the mainstream tokenizer to split a text into a sequence of tokens (e.g., subwords/words/phrases). Thus, enhancing BPE could boost the performance of LLMs and have positive implications for various applications.

\subsection{Language Models}
Language models are designed to maximize the likelihood of a token sequence. Following GPT-3 \citep{brown2020language}, which features 175 billion parameters and demonstrates versatility across a wide range of applications, there has been a significant push towards developing large generative language models like Gopher \citep{rae2021scaling}, PaLM \citep{chowdhery2023palm}, GaLM \citep{du2022glam}, OPT \citep{zhang2022opt}, and LLaMA \citep{touvron2023llama}. Such a surge in development has greatly advanced the fields of natural language understanding and generation.


\subsection{Byte Pair Encoding}

Early neural models had difficulty managing rare words due to limited vocabulary sizes. BPE \cite{sennrich2015neural} effectively addresses that by generating a subword vocabulary. Initially, a corpus is split into characters or bytes, which act as initial tokens. The algorithm iteratively finds the most frequent token pair in the sequence, merges them into a new token, and adds it to the vocabulary until it reaches a predetermined size. The vocabulary is then utilized during the encoding phase to represent any text. Recent advancements like BPE-dropout \citep{provilkov2019bpe} and optimal vocabulary size search \citep{xu2020vocabulary,gowda2020finding,salesky2020optimizing} continue to enrich BPE developments.

However, previous works did not take into account a fundamental flaw of BPE: the iterative training process of BPE incorporates low-frequency tokens into the vocabulary, hindering the inclusion of other actual high-frequency tokens, thus resulting in an imbalance of token frequencies and wastage of the vocabulary. To address that issue, here we propose Scaffold-BPE, which has a dynamic scaffold token removal mechanism that ensures the tokens fed into models are always actual high-frequency tokens.

\section{Methodology}

To enhance the original BPE, we propose Scaffold-BPE to remove the scaffold tokens introduced by the original BPE. Our Scaffold-BPE is simple yet effective. 
In the training process, the Scaffold-BPE dynamically marks scaffold tokens in the vocabulary at each iteration, and finally yields an expanded vocabulary consisting of both normal tokens with the amount equaling the predetermined vocabulary size and several scaffold tokens. In the encoding process, apart from using the normal tokens, Scaffold-BPE temporarily uses scaffold tokens as intermediate tokens to merge into longer normal tokens.

\subsection{Training Process}

The original BPE is trained on a text corpus $C$ with a predefined vocabulary size $N$. After training, BPE returns a vocabulary $V$ consisting of $N$ tokens. For simplicity, $C$ is firstly split into a sequence of smallest unit tokens (denoted as $L$), with each token being a single character/byte. We define $a$, $b$ as two tokens, $(a,b)$ as a token pair, and $f(\cdot)$ as the frequency of a token or token pair within $L$. BPE is trained iteratively. In each iteration, BPE identifies the token pair with the highest frequency: 
\begin{equation}
(a,b) = \arg\max_{(x,y) \in L} f((x,y))\label{ab}
\end{equation} 
BPE then merges (i.e., concatenates) them into a new token $t$, and includes $t$ in $V$. Then BPE updates $L$ via replacing all $(a, b)$ with $t$, and restarts the process again. 

The iterative process of identifying the most frequent token pair $(a, b)$ can be accelerated using a priority queue $Q$. At the beginning of the training process, all token pairs in $L$ are pushed into $Q$ with a descending order of frequency. And after the token pair $(a, b)$ is merged into $t$ in each iteration, BPE updates the frequencies and the ranks of token pairs related to all indexed occurrences of $(a, b)$. For instance, given $(a, b)$ in a context of ``$\ldots, u, a, b, v, \ldots$" in $L$, when $(a, b)$ is replaced with $t$, the frequency of $(u, a)$ or $(b, v)$ would decrease by 1, and meanwhile that of $(u, t)$ or $(t, v)$ would increase by 1. With the occurrences of all token pairs being indexed, there is no need to scan $L$ again and re-count the frequencies of all candidate token pairs for a new iteration. After updating the adjacent token pairs related to $(a, b)$ (i.e, $t$), the frequencies of token pairs like $(u, a)$ or $(b, v)$ would be updated in $Q$, and meanwhile the new candidate token pairs $(u, t)$ and $(t, v)$ would also be pushed into $Q$ with their frequencies.

\begin{algorithm}[t]
\caption{Scaffold-BPE Training Algorithm}
\begin{algorithmic}[1]
\REQUIRE Text Corpus $C$, Vocabulary Size $N$
\STATE Initialize an expanded vocabulary $E$, consisting of a normal-token vocabulary $V$ and a scaffold-token vocabulary $S$
\STATE Split $C$ into a characters/bytes list $L$
\STATE Initialize a priority queue $Q$ storing token pairs within $L$, arranged in reverse order of frequency
\WHILE{$|V| < N$}
    \STATE $(a,b)$ $\leftarrow$ pop $Q_{head}$
    \STATE Merge pair $(a,b)$ into a new token $t$ 
    
    \IF {$t$ in $S$} 
        \STATE /* $t$ may be a previously marked scaffold token */
        \STATE Scaffold$(t)\leftarrow$ False
        \STATE \textbf{continue}
    \ENDIF 
    \STATE Add $t$ to $E$ as a normal token
    \STATE Replace all of $(a,b)$ in $L$ with $t$
    \STATE Update $Q$ \\
    
    \STATE /** \textbf{Scaffold-BPE} Modification Begins **/
    \FOR{each $t'$ in \{$a$, $b$\}}
        \STATE $f(t') \leftarrow f(t') - f(t)$
        \IF {$t'$ in $V$ and $f(t') < f(Q_{head})$}
            \STATE Scaffold$(t')\leftarrow$ True
            \STATE Push $t'$ back to $Q$
        \ENDIF
    \ENDFOR
    \STATE /** \textbf{Scaffold-BPE} Modification Ends **/ \\
    
\ENDWHILE
\RETURN $E$ (with $V$ and $S$ both included)
\end{algorithmic}
\label{Training Algorithm of Enhanced Byte-Pair Encoding}
\end{algorithm}

The Scaffold-BPE expands the vocabulary $V$ to an expanded vocabulary $E$, and assigns an attribute (denoted as $\text{Scaffold}(\cdot)$) to each token in the vocabulary indicating whether it is a scaffold token or not. Thus, the expanded vocabulary $E$ comprises two types of tokens, i.e., normal ones and scaffold ones. We denote all the non-scaffold tokens by $V$, which, as with the original BPE, are the tokens actually used in representing texts for NLP model training:
\begin{equation}
V=\{t \in E \mid \text{Scaffold}(t)=False\}
\end{equation}
Additionally, we denote all the scaffold tokens by $S$, which are not fed into the model, nor do they appear in any token representations after encoding:
\begin{equation}
S=\{t \in E \mid \text{Scaffold}(t)=True\}
\end{equation}
They only serve as intermediate tokens to aid in the training and encoding processes of Scaffold-BPE. Therefore, when calculating vocabulary size, the count of scaffold tokens is not included; only the number of tokens in $V$ is considered.

Initially, a token pair is merged and added to $E$ due to its high frequency. Similarly, Scaffold-BPE marks a token as a scaffold token when its frequency decreases too much. Throughout the entire training process of BPE, $f(a)$ and $f(b)$ only decrease when the token pair $(a,b)$ is merged into a new token $t$. Therefore, as presented in Algorithm \ref{Training Algorithm of Enhanced Byte-Pair Encoding}, Scaffold-BPE introduces an additional step at the end of each iteration, utilizing the decreased $f(a)$ and $f(b)$ to evaluate whether $a$ and $b$ remain high-frequency. If they are no longer considered high-frequency, they would be marked as scaffold tokens. 
Naturally, the token pair at the head of the priority queue $Q$ (denoted as $Q_{head}$) is the next candidate to be added to the vocabulary. Then $f(Q_{head})$ is a natural frequency delimiter between in-vocabulary and out-of-vocabulary tokens. Therefore, if $f(a)$ (or $f(b)$) $< f(Q_{head})$, $a$ (or $b$) is marked as a scaffold token, which means it is not included in $V$:
\begin{equation}
\text{Scaffold}(a) = 
\begin{cases} 
\text{True}, & \text{if } f(a) < f(Q_{head}) \\
\text{False}, & \text{otherwise}
\end{cases}
\end{equation}
Notably, such an additional step leverages the inherent mechanism of BPE without introducing any additional hyper-parameters, maintaining the simplicity and clarity of BPE. Moreover, $f(Q_{head})$ is dynamically adjusted in each iteration, ensuring that Scaffold-BPE can adaptively identify scaffold tokens at any iteration step. Furthermore, scaffold tokens are not permanently marked. They are pushed back into $Q$, reserving the possibility of being ranked top at the priority queue and re-integrated into $V$ in a future iteration.

\begin{algorithm}[t]
\caption{Scaffold-BPE Encoding Algorithm}
\label{Encoding Algorithm of Byte-Pair Encoding}
\begin{algorithmic}[1]
\REQUIRE A Text $T$, Expanded Vocabulary $E$
\STATE Split $T$ into a character/byte token representation (denoted as $R$)
\WHILE{True}
    \STATE /******** \textbf{Scaffolding} Begins ********/
    \STATE Identify all possible merges $M$ using $E$, ignoring token types 
    \STATE /********* \textbf{Scaffolding} Ends *********/ \\
    \IF{$M$ is empty}
        \STATE \textbf{break}
    \ENDIF
    \STATE Select $m$ which is ranked before the others in $E$ from $M$
    \STATE Apply $m$ to $R$
\ENDWHILE \\
\STATE /******** \textbf{Demolishing} Begins ********/
\STATE Demolish all scaffold tokens in $R$ into its shortest non-scaffold child token sequence
\STATE /********* \textbf{Demolishing} Ends *********/ \\
\RETURN $R$
\end{algorithmic}
\end{algorithm}

\subsection{Encoding Process}

The encoding process of the original BPE encodes a text $T$ into a token representation (i.e., $R$) using the vocabulary $V$ generated by BPE training. Firstly, $R$ is a sequence of smallest unit tokens (i.e., character/byte tokens), obtained by splitting $T$. And then, following the ranks of tokens in $V$ as merging priority (i.e., tokens added earlier have higher frequency and thus are assigned higher priority to be merged into), token pairs in $R$ are iteratively merged to build the final representation. 

Similarly, the modifications of Scaffold-BPE in the encoding process are straightforward. Compared to the original BPE, the expanded vocabulary $E$ is utilized. In each iteration, the token $t$ to be merged would be selected from both normal tokens and scaffold tokens:
\begin{equation}
t = \arg\min_{t \in E} rank_E(t)
\end{equation}
where $rank_E(\cdot)$ denotes the rank of a token in $E$. Consequently, during the encoding process, the count of different tokens used actually exceeds the predefined vocabulary size (i.e., $N$). And scaffold tokens are employed as intermediate tokens to merge into longer tokens. We term such a mechanism as \textbf{Scaffolding}, as shown in Algorithm \ref{Encoding Algorithm of Byte-Pair Encoding}.

\begin{figure}[t]
\centering
\includegraphics[width=0.95\columnwidth]{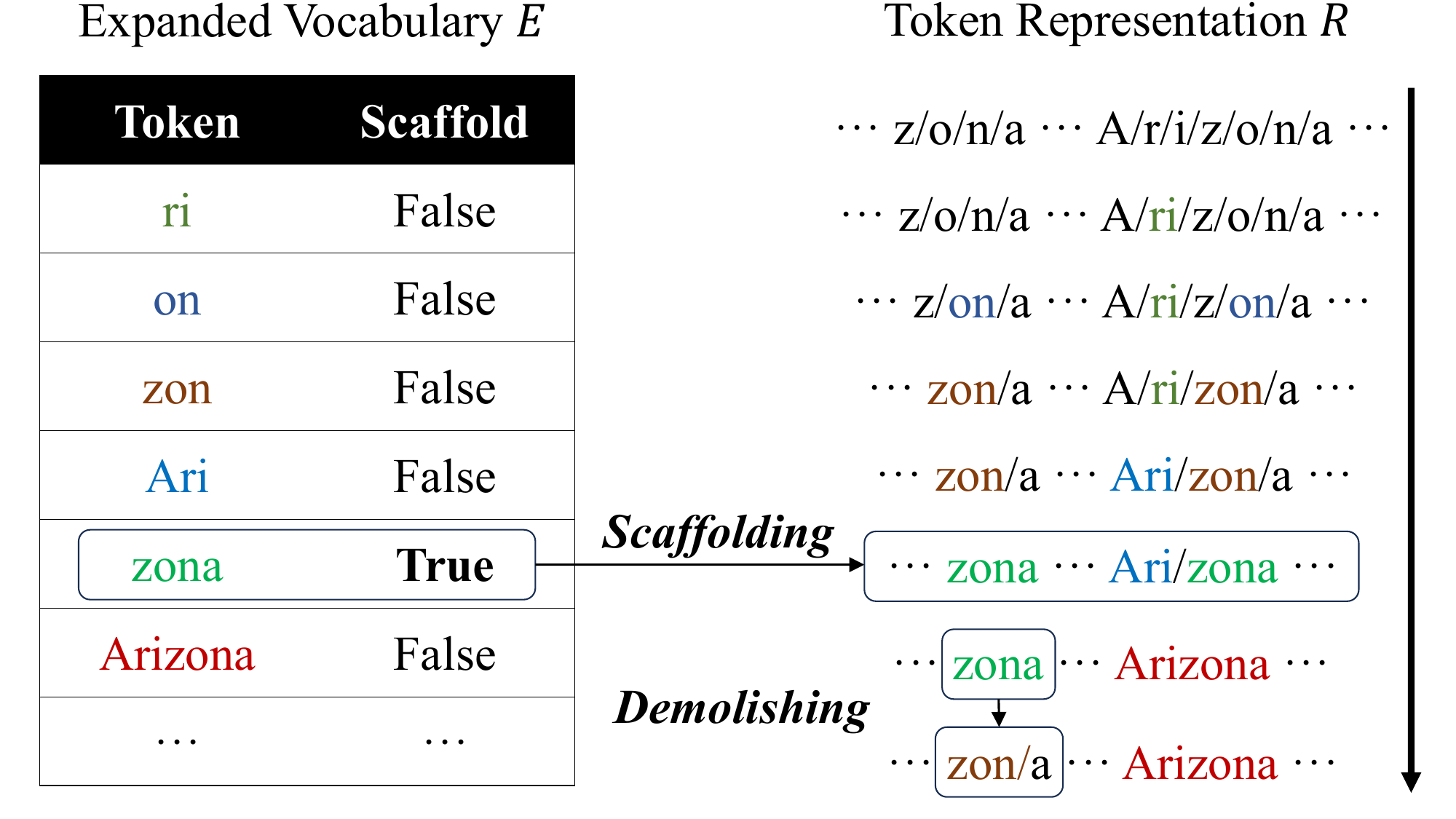}
\caption{The encoding process (i.e., Scaffolding and Demolishing) for the word ``zona" and ``Arizona".}
\label{encoding}
\end{figure}

When no more token pairs can be merged in $R$, the original BPE returns $R$ as the final result. However, due to the introduction of the Scaffolding mechanism in Scaffold-BPE, $R$ may contain scaffold tokens from $S$, potentially increasing the variety of tokens beyond the predefined vocabulary size and exceeding the range of word embeddings that the model can map. To address it, Scaffold-BPE adds one additional step termed as \textbf{Demolishing} at the end of the encoding process. Scaffold-BPE demolishes all scaffold tokens in $R$ into their shortest non-scaffold child token sequences, ensuring that $R$ only consists of tokens from $V$. For example, as shown in Figure \ref{encoding}, the remaining ``zona'' in $R$ is demolished into ``zon'' and ``a''. 
The demolishing step can be formulated as follows:
\begin{equation}
t = 
\begin{cases} 
t, & \text{if Scaffold}(t) = \text{False} \\
(a, b), & \text{otherwise}
\end{cases}
\end{equation}
where $a$ and $b$ are the components of the scaffold token $t$. The formula above would be recursively applied to $a$ and $b$ to derive the shortest non-scaffold child token sequence for $t$.
After the Demolishing step, Scaffold-BPE returns the final token sequence representation (i.e., $R$) for $T$. Since the shortest non-scaffold child token sequences for all scaffold tokens can be precomputed during the training process, the time complexity of demolishing one token is $O(1)$, making its impact on encoding efficiency negligible.

\section{Experiments}
We employ the recently well-attended language modeling tasks to validate the effectiveness of the Scaffold-BPE.

\subsection{Experimental Setup}

\subsubsection{Datasets.}
Our models are trained on the Pile \citep{gao2020pile} dataset, an 825.18 GiB English text dataset designed for training LLMs. 
The data distribution for our model training is identical to that of the original work \cite{gao2020pile}.

\subsubsection{Tokenizer.}
We train two 32K vocabularies (size applied by LLaMA series \cite{touvron2023llama,touvron2023llama2}) using the original BPE and Scaffold-BPE, respectively. 
Similar to GPT-2 \citep{radford2019language}, pre-tokenization was employed to prevent the merging of tokens from different character categories. And following \cite{touvron2023llama}, we split numbers into individual digits.


\subsubsection{Model.}
\label{model}
We train three language models with 468M, 1.2B, and 6.7B parameters, respectively. Specifically, the architectures of the 468M and the 1.2B models are identical to those of the 410M and the 1.0B models outlined in Pythia \cite{biderman2023pythia}. The minor differences in parameter sizes are attributed to the variations in vocabulary size in the embedding and output layer. As for the 6.7B model, its architecture is identical to LLaMA-7B \cite{touvron2023llama}. 

\subsubsection{Training.}
Following the pretraining settings of previous works \cite{xie2023doremi,su2024maskmoe,xiong2024temporal} and limited by our computation budget, by default all models are pretrained with 100B tokens. Note that the volume of corresponding text data contained in an equal amount of tokens is slightly different between the two tokenizers. Considering model training efficiency and commonly used criteria (i.e., the token amount) of computation budget in LLM training, we compare experiments in the setting of an equal amount of training tokens. In the Section \ref{EqualText}, we further analyze both tokenizers in the setting of an equal amount of training text volume.


\subsection{Experimental Results}
\label{Experimental Results}

\begin{table*}[t]
\centering
\resizebox{\linewidth}{!}{
\begin{tabular}
{cc|ccccccccccc}
\toprule
& & \textbf{BoolQ} & \textbf{HellaSwag} & \textbf{OpenBookQA} & \textbf{PIQA} & \textbf{SIQA} & \textbf{StoryCloze} & \textbf{Winogrande} \\
\midrule
\multirow{2}{*}{468M} & Original BPE & 58.64 & 40.78 & 30.50 & 66.57 & 43.40 & 62.77 & 53.00 \\
& Scaffold-BPE & \underline{\textbf{60.52}} & \underline{\textbf{41.68}} & \underline{\textbf{32.20}} & \underline{\textbf{68.69}} & \underline{\textbf{44.09}} & \underline{\textbf{63.04}} & \underline{\textbf{54.22}} \\
\midrule
\multirow{2}{*}{1.2B} & Original BPE & 60.86 & 47.25 & 31.70 & 68.55 & 44.09 & 65.61 & 55.52 \\
& Scaffold-BPE & \underline{\textbf{62.26}} & \underline{\textbf{48.07}} & \underline{\textbf{32.90}} & \underline{\textbf{69.86}} & \underline{\textbf{45.34}} & \underline{\textbf{67.02}} & \underline{\textbf{56.00}} \\
\midrule
\multirow{2}{*}{6.7B} & Original BPE & 62.87 & 60.57 & 35.10 & 73.69 & 46.98 & 71.43 & 60.97 \\
& Scaffold-BPE & \underline{\textbf{64.95}} & \underline{\textbf{61.19}} & \underline{\textbf{38.00}} & \underline{\textbf{74.54}} & \underline{\textbf{47.49}} & \underline{\textbf{72.26}} & \underline{\textbf{61.76}} \\
\bottomrule
\end{tabular}
}
\caption{
At varying model scales, the average accuracy on 0/5-shot common sense reasoning benchmarks ($p$-value $<0.01$).
}
\label{Language Modeling Results}
\end{table*}


\subsubsection{Common Sense Reasoning.}
Our analysis incorporates 7 benchmarks recognized for evaluating common sense reasoning, including BoolQ \citep{clark2019boolq}, HellaSwag \citep{zellers2019hellaswag}, OpenBookQA \citep{mihaylov2018can}, PIQA \citep{bisk2020piqa}, SIQA \citep{sap2019socialiqa}, StoryCloze \citep{mostafazadeh2016corpus}, and Winogrande \citep{sakaguchi2021winogrande}. We present the performance of all models in terms of average accuracy in 0-shot and 5-shot settings.

As shown in Table \ref{Language Modeling Results}, we can observe that the Scaffold-BPE consistently outperforms the original BPE on different setups with different model sizes. Notably, the 6.7B model trained with Scaffold-BPE can achieve a significant 2.08pp (percent point) improvement on BoolQ and a 2.90pp improvement on OpenBookQA. We conduct a $t$-test, and all metrics have $p$-values less than 0.01, indicating that the results are statistically significant.

Such results clearly demonstrate that although the modifications are simple, our proposed Scaffold-BPE is convincingly effective. We attribute it to that Scaffold-BPE can encode text into tokens with a more balanced frequency distribution, which can help language models to learn all tokens more thoroughly.

\begin{table}[t]
\centering
\begin{tabular}{c|cc}
\toprule
  & \textbf{TriviaQA} & \textbf{WebQuestions} \\
\midrule
Original BPE & 15.63 & 8.56  \\
Scaffold-BPE & \underline{\textbf{18.86}} & \underline{\textbf{9.89}} \\
\midrule
\end{tabular}
\caption{
The average exact match performance on 0/5-shot closed-book question-answering benchmarks of the 6.7B-parameter models ($p$-value $<0.01$).
}
\label{Closed Book}
\end{table}

\begin{figure}[t]
\centering
\includegraphics[width=0.95\columnwidth]{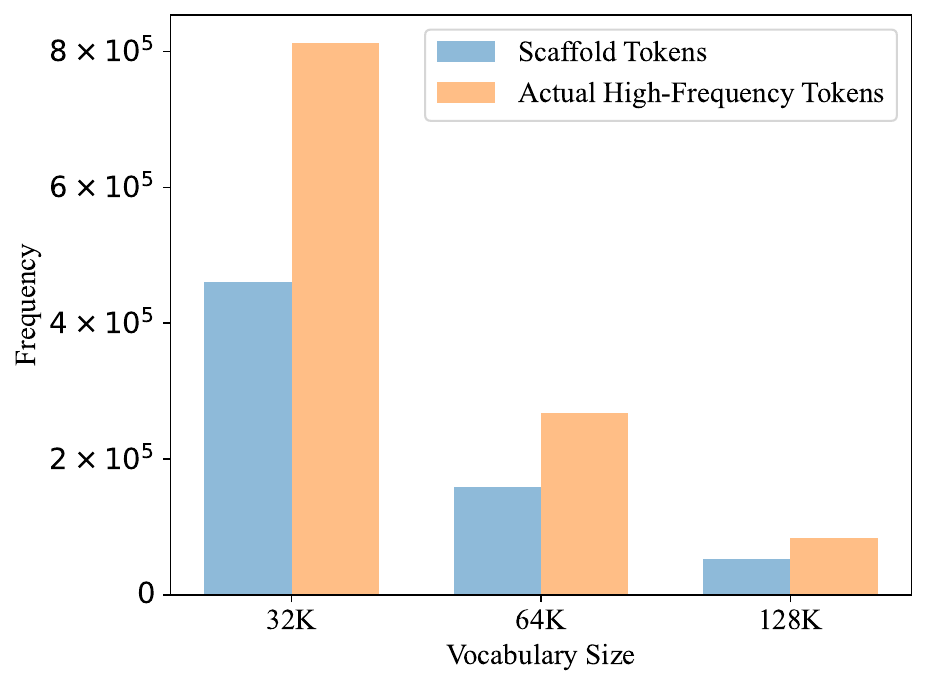}
\caption{The average frequencies of the scaffold tokens and the new actual high-frequency tokens that replace the scaffold tokens in vocabularies of 32K, 64K and 128K.}
\label{tail_mean_token_freq}
\end{figure}

\subsubsection{Closed Book Question Answering.}
For the task of closed book question answering \cite{brown2020language,touvron2023llama}, we evaluate the performance of the largest 6.7B-parameter models with different tokenizers on 2 benchmark datasets, i.e., TriviaQA \citep{joshi2017triviaqa} and WebQuestions \citep{berant2013semantic}. We report the exact match performance for the zero-shot and few-shot settings in Table \ref{Closed Book}. 
It can be seen that the model trained with the proposed Scaffold-BPE can achieve a 3.23pp improvement on TriviaQA and a 1.33pp improvement on WebQuestions, with both $p$-values less than 0.01. All results above demonstrate that Scaffold-BPE can enhance model performance across different types of downstream tasks.

\subsection{Discussion}

\subsubsection{Various Vocabulary Size.}
\label{Robustness In Language Modeling Tasks}
Depending on the size of the training corpus, the diversity of the languages, the size of the model, and the types of tasks, different vocabulary sizes are set in practice. Therefore, to validate the robustness of Scaffold-BPE across various vocabulary sizes, in addition to the 32K vocabulary \citep{touvron2023llama}, we also trained two vocabularies sized at 64K \citep{baichuan7b,baichuan13b} and 128K \citep{yang2023baichuan}.
The experimental setup is identical to that of the 468M-parameter model mentioned before.

As shown in Figure \ref{tail_mean_token_freq}, Scaffold-BPE can replace scaffold tokens in vocabularies with actual high-frequency tokens, which significantly increases the average frequencies of those tokens. The frequency improvements are 76.40\%, 68.58\%, and 58.99\% for the 32K, 64K, and 128K vocabulary sizes, respectively. The enhancement in token frequency distribution effectively promotes the learning of those tokens, which can contribute to better model performance across various tasks.


\begin{table}[t]
\centering
\setlength{\tabcolsep}{0.3mm}
\small
\resizebox{\columnwidth}{!}{
\begin{tabular}{c|cc|cc}
\toprule
& \multicolumn{2}{c|}{\textbf{64K}} & \multicolumn{2}{c}{\textbf{128K}} \\
& \textbf{Orig. BPE} & \textbf{Scaffold-BPE} & \textbf{Orig. BPE} & \textbf{Scaffold-BPE} \\
\midrule
BoolQ & 58.01 & \underline{\textbf{59.71}} & 56.67 & \underline{\textbf{59.43}} \\
HellaSwag & 41.82 & \underline{\textbf{42.06}} & 42.70 & \underline{\textbf{42.91}} \\
OpenBookQA & 30.90 & \underline{\textbf{31.10}} & 31.10 & \underline{\textbf{32.30}} \\
PIQA & 67.95 & \underline{\textbf{69.26}} & 67.68 & \underline{\textbf{68.82}} \\
SIQA & 43.47 & \underline{\textbf{43.86}} & 43.83 & \underline{\textbf{44.14}} \\
StoryCloze & 64.19 & \underline{\textbf{65.02}} & 64.11 & \underline{\textbf{65.26}} \\
Winogrande & 53.67 & \underline{\textbf{54.34}} & 53.91 & \underline{\textbf{55.09}} \\
\bottomrule
\end{tabular}
}
\caption{
At varying vocabulary sizes, the average accuracy on 0/5-shot common sense reasoning benchmarks ($p$-value $<0.01$).
}
\label{Vocab Size Results}
\end{table}

\begin{table}[t]
\centering
\resizebox{\columnwidth}{!}{
\begin{tabular}{c|cc}
\toprule
 & \textbf{Original BPE} & \textbf{Scaffold-BPE} \\
\midrule
BoolQ & 58.21 & \underline{\textbf{61.62}} \\
HellaSwag & 45.10 & \underline{\textbf{46.03}} \\
OpenBookQA & 31.10 & \underline{\textbf{33.50}} \\
PIQA & 68.63 & \underline{\textbf{69.86}} \\
SIQA & 43.78 & \underline{\textbf{44.73}} \\
StoryCloze & 65.53 & \underline{\textbf{66.54}} \\
Winogrande & 53.67 & \underline{\textbf{56.12}} \\
\bottomrule
\end{tabular}
}
\caption{
At 300B training tokens, the average accuracy on 0/5-shot common sense reasoning benchmarks ($p$-value $<0.01$).
}
\label{Tokens Results}
\end{table}

Moreover, as shown in Table \ref{Vocab Size Results}, the results demonstrate that Scaffold-BPE consistently outperforms the original BPE across all vocabulary sizes, which indicates that the superiority of Scaffold-BPE is not sensitive to vocabulary size. Its algorithmic design enables it to adaptively remove scaffold tokens across any vocabulary size, without the need for manually designed or heavily-tuned hyperparameters.

\subsubsection{More Training Tokens.}
According to the Scaling Law, the loss scales as a power-law with model size, dataset size, and the amount of training computation \cite{kaplan2020scaling}. To demonstrate the effectiveness of our Scaffold-BPE with more training tokens, we continue training the 468M models up to 300B tokens \cite{zhang2022opt,biderman2023pythia}. 

As shown in Table \ref{Tokens Results}, the results demonstrate that Scaffold-BPE consistently outperforms the original BPE at 300B training tokens, well indicating that in the era of increasingly large training datasets for LLMs, our Scaffold-BPE can effectively enhance the capabilities of those models through simple modifications to the original BPE.

\subsubsection{Applicable for Other Tasks, Languages, Model Architectures and Compatible with Other BPE Enhancements.}
Although the development of LLMs is burgeoning, some applications still prefer using conventional models due to their lower training and inference costs. In the NLP field, BPE was initially combined with transformer models and applied to machine translation tasks \cite{sennrich2015neural}, which typically face an open vocabulary challenge and involve substantial textual variations between two languages. 
Therefore, to validate the versatility of the Scaffold-BPE method, we additionally conduct evaluations on machine translation tasks with identical experimental setup on WMT'14 En-De and En-Fr dataset in the prior work \cite{ott-etal-2018-scaling}.

As shown in Table \ref{Machine Translation Results}, Scaffold-BPE outperforms the original BPE in machine translation tasks, which demonstrates that Scaffold-BPE is not specific to language modeling tasks and can be applied to a wider range of tasks.

Besides, experiments conducted with En-De and En-Fr language pairs demonstrate that Scaffold-BPE is language insensitive. Scaffold-BPE is capable of identifying and removing the scaffold tokens introduced by the original BPE across different languages.

Moreover, previous experiments on language modeling tasks are carried out on the decoder-only architecture. For the machine translation tasks, we utilize the encoder-decoder architecture \cite{vaswani2017attention}. The exceptional performance of Scaffold-BPE confirms its architecture insensitivity, indicating its applicability across a wider range of neural network architectures.

Finally, Scaffold-BPE is orthogonal to and can be combined with existing enhancements to BPE, like BPE-Dropout \cite{provilkov2019bpe}. As shown in Table \ref{Machine Translation Results}, Scaffold-BPE with BPE-Dropout achieves further improvements on BLEU, well indicating the compatibility of Scaffold-BPE.

\begin{table}[t]
\centering
\begin{tabular}{c|cc}
\toprule
\textbf{} & \textbf{En-De} & \textbf{En-Fr} \\
\midrule
Original BPE & 29.31 & 43.20 \\
\quad\quad\quad\quad + BPE-Dropout & 29.50 & 43.44 \\
Scaffold-BPE & \underline{\textbf{29.76}} & \underline{\textbf{43.81}} \\
\quad\quad\quad\quad + BPE-Dropout & \underline{\textbf{29.78}} & \underline{\textbf{43.83}} \\
\bottomrule
\end{tabular}
\caption{
BLEU on WMT'14 En–De and En–Fr.
}
\label{Machine Translation Results}
\end{table}

\begin{table}[t]
\centering
\begin{tabular}{c|ccccccc}
\toprule
 & \textbf{Pile}  & \textbf{En-De} & \textbf{En-Fr}\\
\midrule
Original BPE & 3.879 & 4.830 & 5.012 \\
Scaffold-BPE & \underline{\textbf{3.889}} & \underline{\textbf{4.861}} & \underline{\textbf{5.042}} \\
\bottomrule
\end{tabular}
\caption{
Compression Rate (the average number of bytes per token) on the Pile dataset and the WMT dataset.
}
\label{Compression Rate}
\end{table}

\subsubsection{Higher Compression Rate.}
Besides the performance of models on downstream NLP tasks, the compression rate for a given text corpus is a metric to measure the effectiveness of a tokenizer. A higher compression rate means that fewer tokens are required to represent the same corpus. As shown in Table \ref{Compression Rate}, Scaffold-BPE, utilizing a scaffold tokens removal mechanism, retains more actual high-frequency tokens in the final vocabulary, and thus it achieves a higher compression rate on all the corpus in our experiments.

\subsubsection{Experiments under Same Corpus Size}
\label{EqualText}
As mentioned before, considering model training efficiency and commonly used criteria (i.e., the token amount) of computation budget in LLM training, experiments above are compared in the setting of an equal amount of training tokens. To eliminate the impact of different amounts of training text caused by different compression rates on experiment results, we additionally train two 468M-parameter models on exactly 388 GiB training text ($\approx$ 100B tokens). As shown in Table \ref{Text Results}, Scaffold-BPE consistently outperforms the original BPE, demonstrating that the effectiveness of Scaffold-BPE is not merely obtained by allowing models to digest more data in the same computation budget. Our Scaffold-BPE also alleviates the issue of token frequency imbalance, allowing models to learn all tokens more sufficiently and evenly, thus achieving better performance.

\begin{table}[t]
\centering
\resizebox{\columnwidth}{!}{
\begin{tabular}
{c|cccccccc}
\toprule
 & \textbf{Original BPE} & \textbf{Scaffold-BPE} \\
\midrule
BoolQ & 58.72 & \underline{\textbf{60.55}} \\
HellaSwag & 40.84 & \underline{\textbf{41.69}} \\
OpenBookQA & 30.55 & \underline{\textbf{32.22}} \\
PIQA & 66.58 & \underline{\textbf{68.78}} \\
SIQA & 43.40 & \underline{\textbf{44.13}} \\
StoryCloze & 62.85 & \underline{\textbf{63.08}} \\
Winogrande & 53.07 & \underline{\textbf{54.25}} \\
\bottomrule
\end{tabular}
}
\caption{
At exactly 388 GiB training text, the average accuracy on 0/5-shot common sense reasoning benchmarks ($p$-value $<0.01$).
}
\label{Text Results}
\end{table}

\section{Conclusions}
In this paper, we present our observation of tokens with imbalanced frequencies in BPE vocabulary, which we term scaffold tokens. Those scaffold tokens, while integral to the formation of longer tokens, do not represent actual frequent tokens and affect the performance of LLMs negatively. To address that, we propose Scaffold-BPE, which can remove scaffold tokens from the final token representations by dynamically marking scaffold tokens in the training process and temporarily utilizing them in the encoding process. The Scaffold-BPE is parameter-free, computation-light, easy-to-implement, and widely effective, well preserving the simplicity and clarity of BPE. Through extensive experiments, including varying model sizes, vocabulary sizes and more training tokens, etc., Scaffold-BPE demonstrates its robustness and superiority over the original BPE.

\bibliography{acl_latex}
\clearpage

\appendix

\section{Additional Discussion}

\subsection{Higher Entropy, Lower Redundancy}
Scaffold-BPE can alleviate the imbalance in token frequency, which can lead to an increase in information entropy.
We measure Shannon Entropy and Redundancy \cite{gutierrez2021characters} over token representations of texts obtained with the original BPE and our Scaffold-BPE. Both take as input a text $T$ with a vocabulary of (normal) tokens $V=\{t_1, t_2, ..., t_V\}$ of size $|V|$.

Entropy $H$ is a measure of the average information. Where the probability of a token $p(t)$ is estimated using the so-called maximum likelihood method (i.e., its relative frequency in the text). Higher values of Entropy indicate higher complexity (less predictability).
\begin{equation}H(T)=-\sum_{i=1}^Vp(t_i)\log_2p(t_i)\end{equation}
The Redundancy $R$ quantifies how close the empirically estimated entropy is to the maximum value it can take. 
\begin{equation}R(T)=1-\frac{H(T)}{max\{H(T)\}}=1-\frac{H(T)}{\log_2|V|}\end{equation}

As shown in Table \ref{Entropy and Redundancy results}, taking the 32K vocabulary as an example, our Scaffold-BPE can encode Pile dataset \cite{gao2020pile} with higher Entropy and lower Redundancy. Consequently, tokens in the vocabulary of our Scaffold-BPE have more balanced appearing probabilities. According to \citet{su2023infoentropy}, our vocabulary with balanced token occurrences mitigates the learning imbalance problem, resulting in more sufficient learning towards the text corpus, thus achieving better performance.

\subsection{Better Uniformity of Learned Embeddings}
Prior works have analyzed the embedding space learned by a model \cite{provilkov2019bpe} and found that better uniformity prefers a token embedding space that preserves maximal information \cite{wang2020understanding}. To demonstrate our Scaffold-BPE can mitigate token frequency distribution imbalance, thus leading to a better-learned token embedding space with better uniformity, we visualize the token embeddings in the 6.7B-parameter models, following \citet{provilkov2019bpe}. 

\begin{figure}[t]
    \centering
    \subfigure[The Original BPE]
    {
        \begin{minipage}[b]{0.8\linewidth}
            \centering
            \includegraphics[width=\linewidth]{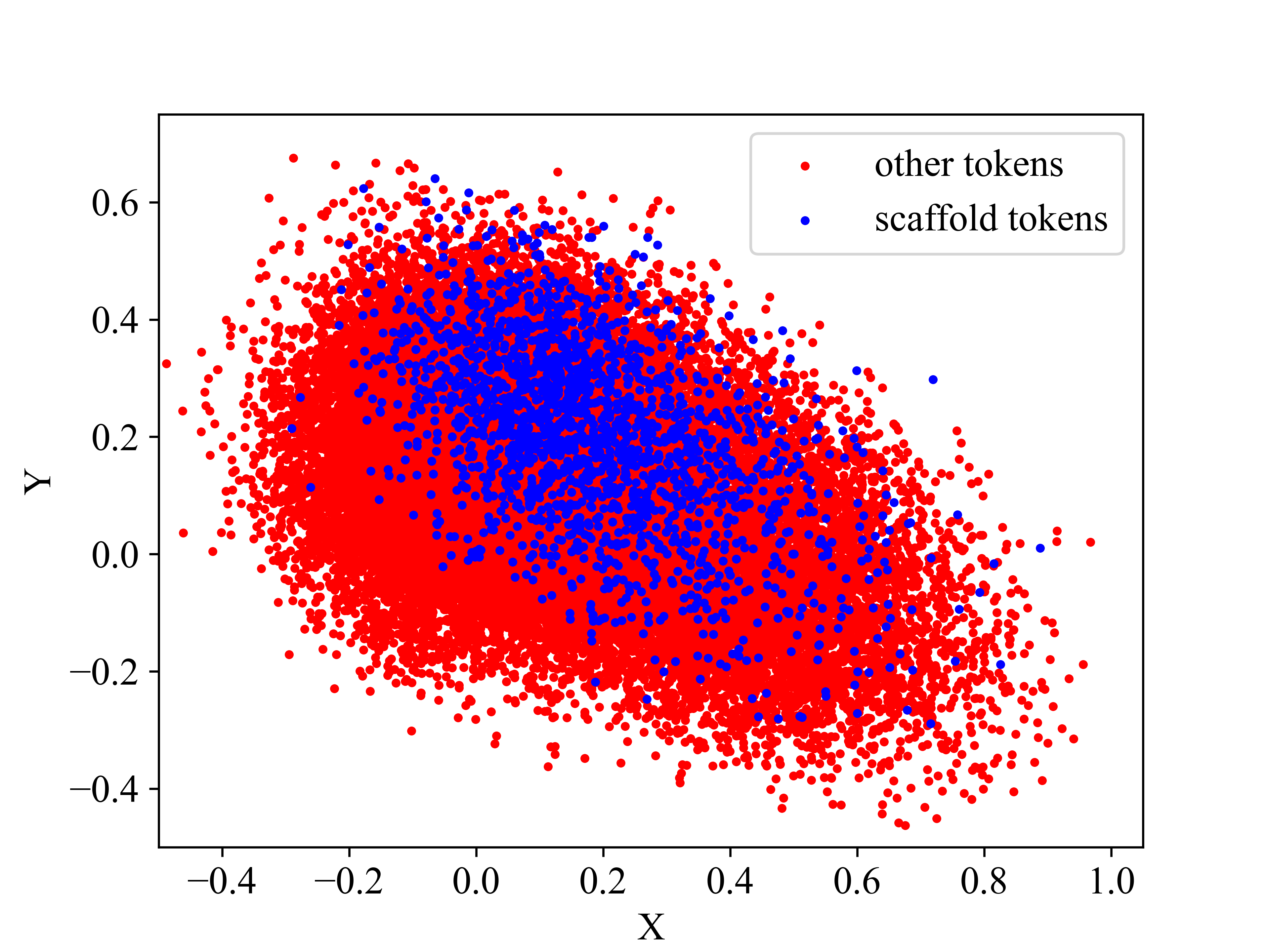}
        \end{minipage}
        \label{encoding-a}
    }
    \subfigure[The Scaffold-BPE]
    {
        \begin{minipage}[b]{0.8\linewidth}
            \centering
            \includegraphics[width=\linewidth]{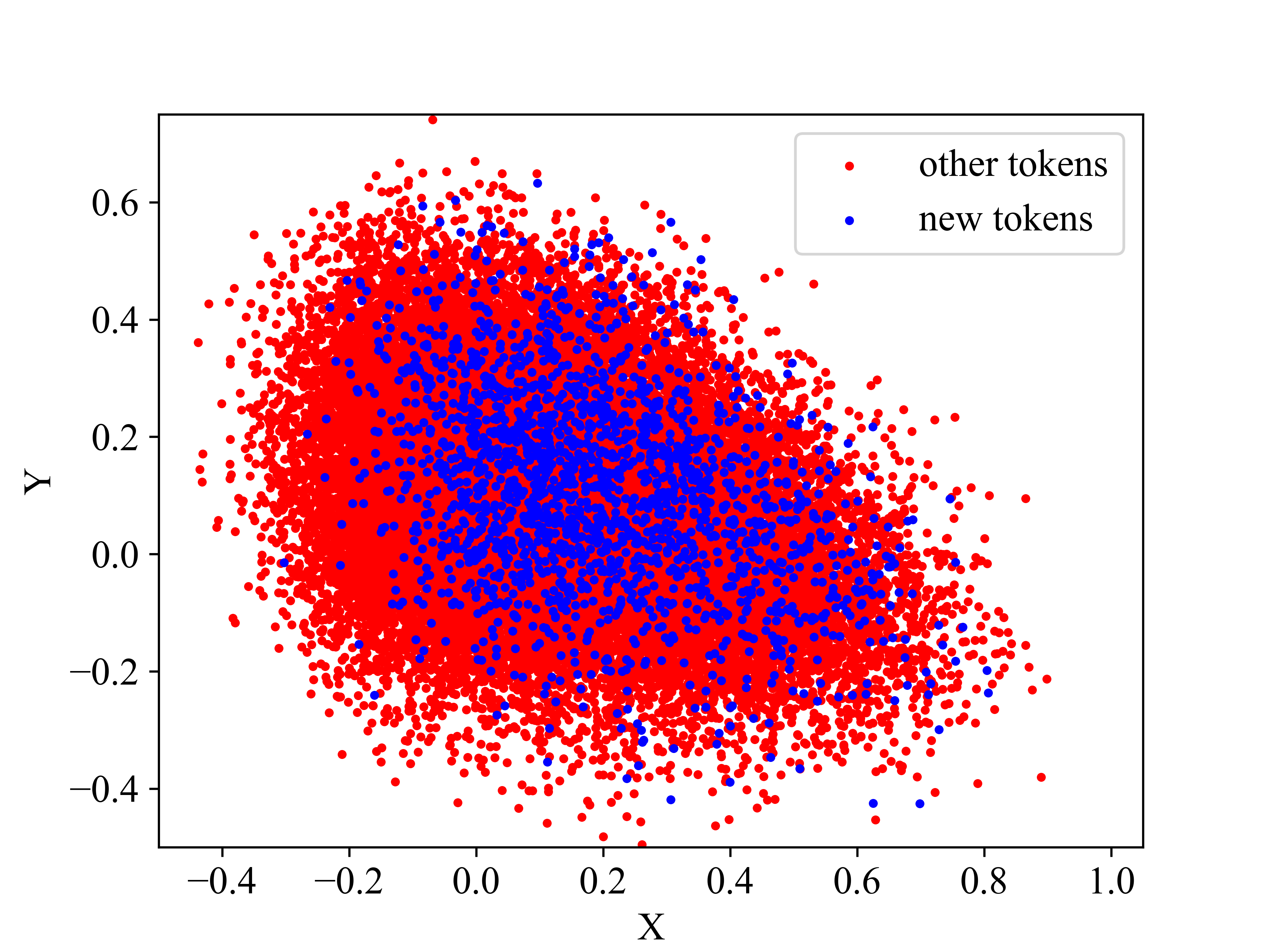}
        \end{minipage}
        \label{encoding-b}
    }
    \caption{Visualization of the learned embeddings of tokens in the respective vocabulary generated by the original BPE and Scaffold-BPE.}
    \label{svd}
\end{figure}

As shown in Figure \ref{svd}, the embeddings of scaffold tokens learned via the original BPE are more clustered, which means they are not well learned. On the contrary, the embeddings of new tokens introduced by Scaffold-BPE after removing scaffold tokens have better uniformity, which are more evenly distributed across the semantic space. Therefore, models trained with Scaffold-BPE can achieve better performance.

\begin{table}[t]
\centering
\begin{tabular}{ccc}
\toprule
\textbf{} & \textbf{Entropy$\uparrow$} & \textbf{Redundancy$\downarrow$} \\
\midrule
Original BPE & 11.2382 & 0.2491 \\
Scaffold-BPE & \underline{\textbf{11.2443}} & \underline{\textbf{0.2487}} \\
\bottomrule
\end{tabular}
\caption{
Entropy and Redundancy on tokenized Pile dataset.
}
\label{Entropy and Redundancy results}
\end{table}

\begin{table}[t]
\centering
\resizebox{\columnwidth}{!}{
\begin{tabular}{cccccc}
\toprule
model size & dimension & $n$ heads & $n$ layers & batch size & seq length \\
\midrule
468M & 1024 & 16 & 24 & 1024 & 1024 \\
1.2B & 2048 & 8 & 16 & 2048 & 1024 \\
6.7B & 4096 & 32 & 32 & 2048 & 2048 \\
\bottomrule
\end{tabular}
}
\caption{
Model sizes and architectures.
}
\label{Model_sizes}
\end{table}

\section{Further Experiment Details}


\subsection{Model Training Hyper-Parameters}

We train three language models with 468M, 1.2B, and 6.7B parameters, respectively. Specifically, the architectures are listed in Table \ref{Model_sizes}. Following LLaMA \citep{touvron2023llama}, we use the AdamW optimizer \citep{loshchilov2017decoupled} with a learning rate of $3.0\times10^{-4}$, $2k$ warmup steps, and a cosine learning rate decay schedule.

\subsection{Evaluation}
For fair comparisons, we utilize the open-source pipeline \texttt{lm-evaluation-harness}
\citep{eval-harness} for evaluation. 
We reported the average evaluation results of the last five checkpoints.

\subsection{Computing Infrastructure}
The experiments were conducted using a cluster of 16 servers, each equipped with 8 NVIDIA H800 GPUs. The total memory available on each server is 600 GB, and each server is powered by a 70-core CPU. The operating system running on these servers is Ubuntu 20.04.6 LTS. We report the version numbers of used packages in Table \ref{tab:packages}.

\begin{table}[t]
\centering
\begin{tabular}{l|l}
\toprule
Package & Version \\
\midrule
absl-py & 1.4.0 \\
accelerate & 0.21.0 \\
datasets & 2.14.3 \\
deepspeed & 0.10.0 \\
ninja & 1.11.1 \\
protobuf & 4.23.4 \\
pytorch-triton & 2.1.0+e6216047b8 \\
scikit-learn & 1.3.0 \\
scipy & 1.11.1 \\
torch & 2.1.0.dev20230807+cu121 \\
torchaudio & 2.1.0.dev20230807+cu121 \\
torchinfo & 1.8.0 \\
torchvision & 0.16.0.dev20230807+cu121 \\
triton & 2.0.0 \\
wandb & 0.15.7 \\
flash-attn & 2.0.4 \\
zstandard & 0.21.0 \\
seaborn & 0.12.2 \\
gradio & 3.39.0 \\
\bottomrule
\end{tabular}
\caption{Versions of used packages.}
\label{tab:packages}
\end{table}

\end{document}